\documentclass[10pt,twocolumn,letterpaper]{article}

\usepackage{cvpr}              % To produce the CAMERA-READY version
\usepackage{amsmath,amssymb}

\usepackage{listings}
\usepackage{xcolor}
\usepackage{enumitem}

\definecolor{cvprblue}{rgb}{0.21,0.49,0.74}
\usepackage[pagebackref,breaklinks,colorlinks,allcolors=cvprblue]{hyperref}

\def\paperID{*****} % *** Enter the Paper ID here
\def\confName{CVPR}
\def\confYear{2025}

\title{2.5D U-Net with Depth Reduction for 3D CryoET Object Identification}

\author{Yusuke Uchida\\
GO Inc.\\
Tokyo, Japan\\
\and
Takaaki Fukui\\
GO Inc.\\
Tokyo, Japan\\
}

\begin{document}
\maketitle
\begin{abstract}
Cryo-electron tomography (cryoET) is a crucial technique for unveiling the structure of protein complexes.  
Automatically analyzing tomograms captured by cryoET is an essential step toward understanding cellular structures.  
In this paper, we introduce the 4th place solution from the CZII - CryoET Object Identification competition, which was organized to advance the development of automated tomogram analysis techniques.
Our solution adopted a heatmap-based keypoint detection approach, utilizing an ensemble of two different types of 2.5D U-Net models with depth reduction.
Despite its highly unified and simple architecture, our method achieved 4th place, demonstrating its effectiveness.

\end{abstract}    
\section{Introduction}
\label{sec:introduction}

Protein complexes (such as oxygen-carrying hemoglobin, or keratin in hair, and thousands of others) are essential for cellular function, and understanding their interactions is crucial for our health and finding new disease treatments. Cryo-electron tomography (cryoET) generates 3D images, known as tomograms, with near-atomic resolution, showing proteins in their very complex and crowded natural environment~\cite{Young2023}. Therefore, cryoET has immense potential to unlock the mysteries of the cell.

Remarkably, a vast amount of data is publicly available in a standardized format through the cryoET data portal~\footnote{\url{https://cryoetdataportal.czscience.com/}}.
However, to fully utilize this data, it is necessary to automatically identify each protein molecule within these cryoET tomograms~\cite{Peck2024}.
To advance automated protein molecule recognition technology, the CZII - CryoET Object Identification competition was launched~\footnote{\url{https://www.kaggle.com/competitions/czii-cryo-et-object-identification}}.

In this competition, seven training samples, each with a size of $630 \times 630 \times 184$, are provided.
Each sample is annotated with the center positions of six types of particles~\cite{Peck2024}.  
Participants are required to predict the centers of these particles.  
The test dataset consists of approximately 500 samples to be predicted.
The predictions are evaluated using the $F_\beta$ metric with $\beta = 4$.
This metric emphasizes recall over precision, imposing heavy penalties on missed particles while being more tolerant of false positives.
Each particle type has a unique distance threshold for correct prediction, determined by its size.
The $F_\beta$ metric is calculated for each particle type, and the final score is obtained by computing the weighted sum using predefined weights based on the difficulty of each type.

This paper presents the detailed approach of the yu4u \& tattaka team, which achieved 4th place in the competition.
Our training source code is publicly available~\footnote{\url{https://github.com/yu4u/kaggle-czii-4th}}\footnote{\url{https://github.com/tattaka/czii-cryo-et-object-identification-public}}.

\section{Proposed Method}

We used a heatmap-based approach to detect particle points, which is the most commonly employed technique in human pose estimation~\cite{Newell2026} and facial keypoint detection~\cite{Bulat2017}.
Since this competition involves 3D images rather than 2D images, we utilized two types of U-Net~\cite{Ronneberger2015} models (yu4u's model and tattaka's model) that take 3D voxels as input and output 3D heatmaps.

\subsection{Validation Strategy}
Designing an appropriate validation strategy is crucial for selecting the model architecture and tuning hyperparameters.  
We adopted a 7-fold cross-validation (CV), where each of the seven training samples was used for validation.
However, since the number of training data was significantly smaller compared to the test data, we observed that the CV scores were not well correlated with the leaderboard scores.

We used the CV scores only to confirm that the metric produced was somewhat reasonable and to select model checkpoints. For models with potential for improvement, we submitted them and relied on the public leaderboard to decide which methods to adopt or discard.

\subsection{Creating the Ground Truth Heatmap}
We generate the ground truth heatmap necessary for model training. This involves converting the ground truth particle coordinates into the pixel coordinate system and creating a mask using a Gaussian function, where the particle center is set to 1.0 and $\sigma$ is 6 pixels for yu4u's model. For tattaka's model, different $\sigma$ values were used for different particles based on their sizes.

We argue that an offset of 1.0 should be added when converting particle coordinates into the pixel coordinate system. While the discussion~\cite{David2025} suggests adding 0.5, we demonstrate that 1.0 is the correct value~\cite{yu4u2025}. The main difference is that the previous discussion assumes the particle center is at the top-left of a pixel, whereas we argue that the circle should be drawn from the pixel center on average.

\subsection{yu4u's Model}
We adopted a 2.5D U-Net~\cite{Kumar2024}, which utilizes a 2D image-based model as the backbone.
The outputs from each stage of this backbone are pooled along the depth direction, enabling hierarchical feature extraction in the depth (Z) dimension as well.
This idea was inspired by the excellent notebook~\cite{hengck232025}.
An interesting observation is that replacing this pooling operation with strided 3D convolutions degrades performance.
This would be because the pooling method effectively aggregates depth features while preserving the original 2D backbone’s feature maps as much as possible.
Similar to many other Kaggle competitions dealing with 3D data, a U-Net utilizing a 2D backbone pretrained with ImageNet outperformed a straightforward U-Net with a 3D backbone~\cite{Cicek2016} in our preliminary experiments.

We also applied 3D convolutions between the encoder and decoder to further extract depth-wise features, inspired by the 3rd place solution of the contrails competition~\cite{knshnb2025}.

Initially, we used a plain 3D U-Net decoder, but processing high-resolution feature maps required significant memory and computation. To address this, we adopted a model that outputs the final heatmap using pixel shuffle from a feature map with a stride of 4. Pixel shuffle~\cite{Shi2016}, also known as \texttt{depth\_to\_space} in TensorFlow, is an operation that redistributes information from the channel dimension to the spatial dimensions. Compared to deconvolution, it offers advantages in computational efficiency and reducing artifacts.
In the other upsampling parts of the decoder, an upsampling layer and 3D convolution blocks are used.

\begin{figure}[tb]
    \centering
    \includegraphics[width=0.5\textwidth]{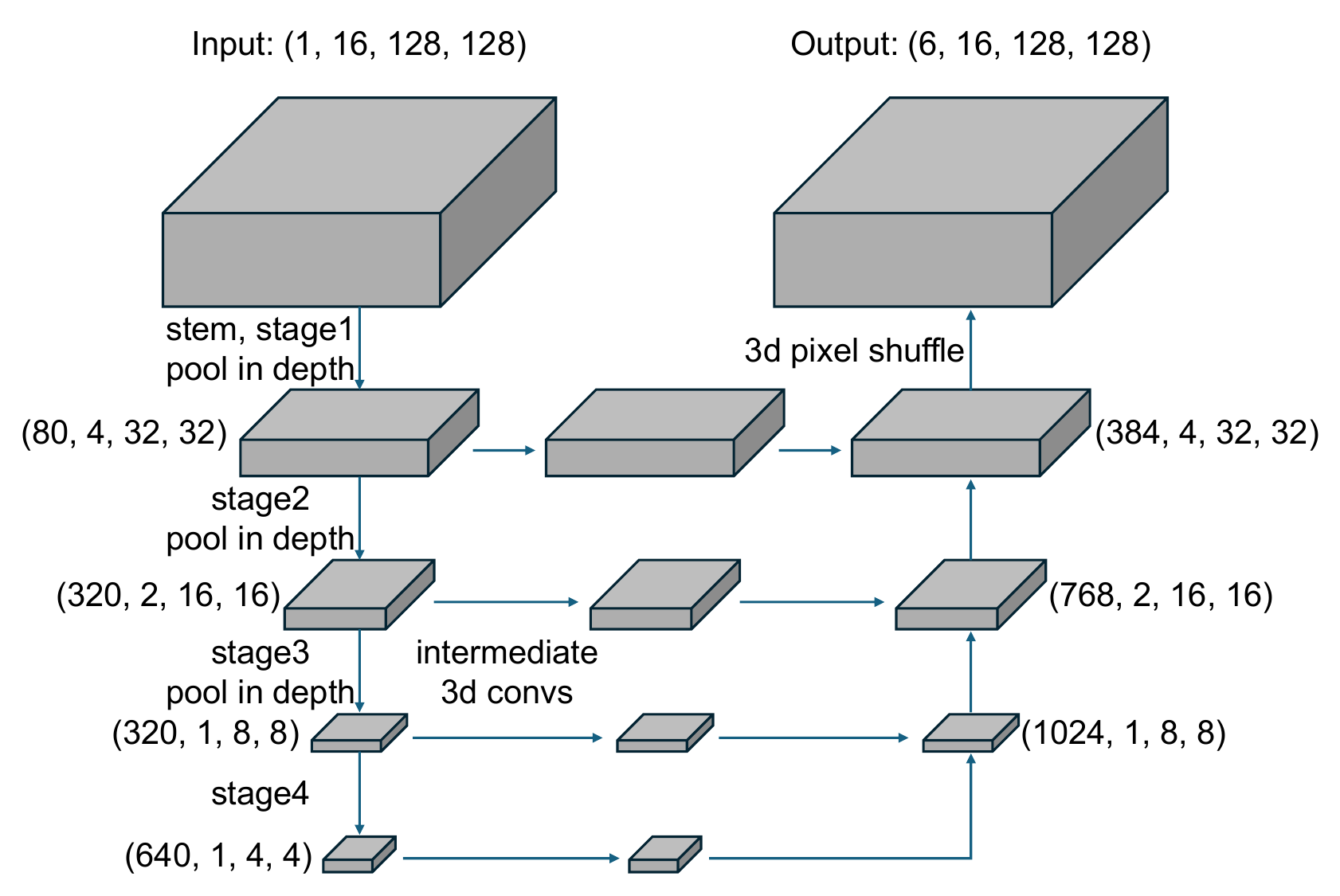}
    \caption{The architecture of yu4u's model.}
    \label{fig:yu4u_model}
\end{figure}

For the final submission, we adopted a ConvNeXt Nano~\cite{Sanghyun2023} model as the backbone.
The overall structure of yu4u's model is shown in Fig.~\ref{fig:yu4u_model}

\begin{table}[tb]
    \centering
    \footnotesize
    \begin{tabular}{ll}
        \toprule
        Training Epochs & 64 \\
        Learning Rate & $10^{-3}$ \\
        Optimizer & AdamW \\
        Weight Decay & 0 \\
        Warmup Epochs & 4 \\
        LR Scheduling Strategy & Cosine Decay \\
        Batch Size & 32 \\
        EMA Decay & 0.999 \\
        \bottomrule
    \end{tabular}
    \caption{Hyperparameters used for training yu4u's model.}
    \label{tab:hyperparams1}
\end{table}

\subsubsection{Loss Function for yu4u's Model}
Since the number of particles within the volume is small, there is a significant class imbalance between positive and negative samples during training.
To address this imbalance problem, we utilized the extended MSE loss function for yu4u's model, where the heatmap values were used as weights.
In practice, since areas without particles would have a weight of zero, a fixed value $\alpha = 0.1$ was added to the heatmap values to be used as weights:
\begin{equation}
    \mathcal{L}_\text{yu4u}(p, y) = \text{mean} (\text{MSE}(p, y) \cdot (y + \alpha)),
\end{equation}
where $p$ is a predicted heatmap and $y$ is the corresponding ground-truth heatmap.
The hyperparameters used for training yu4u's model are shown in Table~\ref{tab:hyperparams1}.

\subsection{tattaka's Model}
This model is a lightweight 2.5D U-Net with ResNetRS50~\cite{Bello2021} as the backbone.
The input to the model is a volume of size $32 \times 128 \times 128$, and it outputs a 3D heatmap of the same size. Within the backbone, the depth is progressively reduced by half using average pooling for the first two stages. After that, average pooling with kernel=3, stride=1, padding=1 is used to maintain the depth while facilitating information exchange along the depth dimension.

\begin{figure}[tb]
    \centering
    \includegraphics[width=0.5\textwidth]{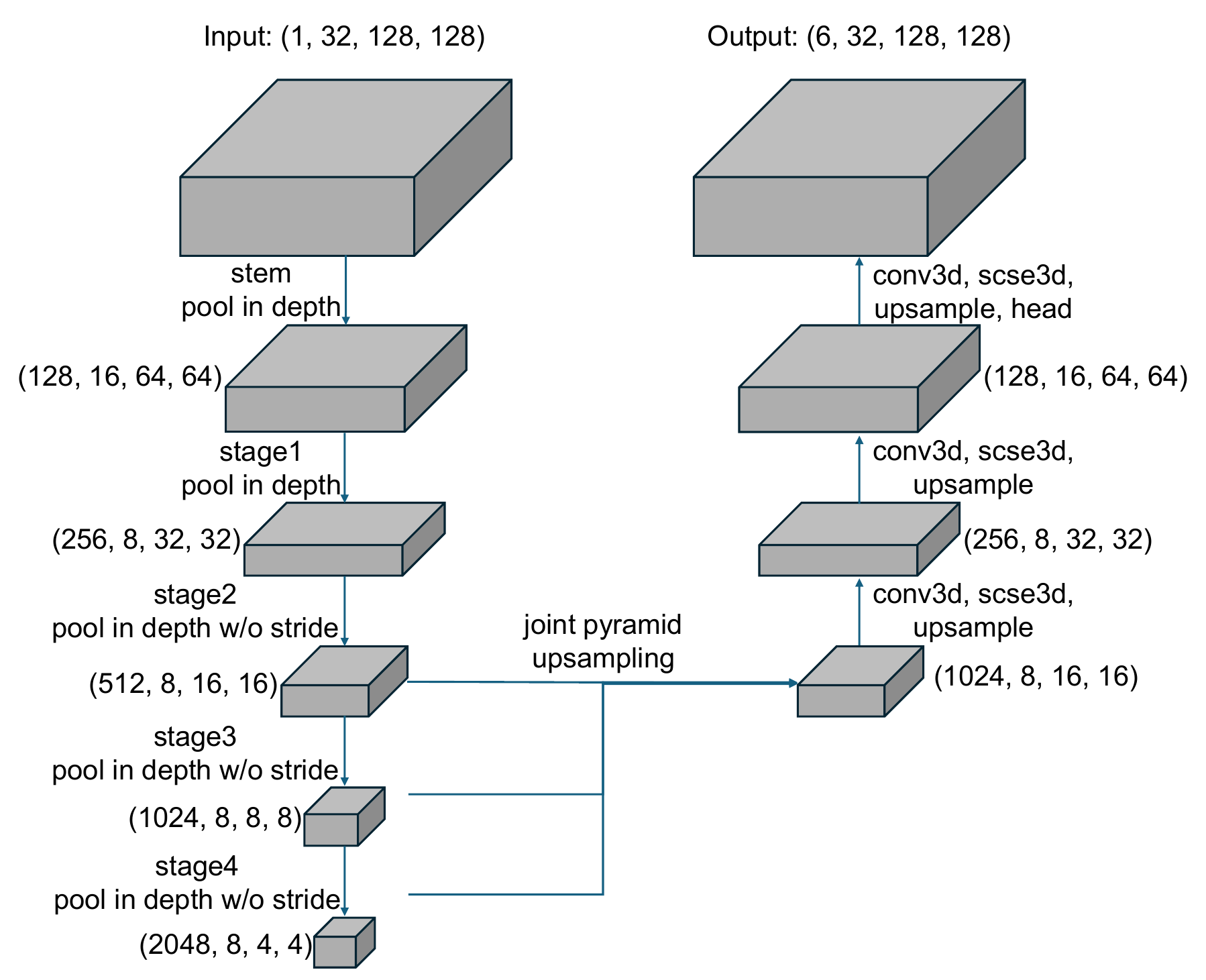}
    \caption{The architecture of tattaka's model.}
    \label{fig:tattaka_model}
\end{figure}

In the decoder, the three lowest-resolution feature maps are fed into joint pyramid upsampling~\cite{wu2019fastfcn} to generate a feature map that contains information at multiple resolutions. The feature map is then progressively upsampled using upsampling blocks until they reach the same size as the input volume.
The upsampling block consists of a 3D conv, an seSC~\cite{Roy2018} attention block, and an upsampling layer.
The overall structure of this 2.5D U-Net is shown in Fig.~\ref{fig:tattaka_model}.

\subsubsection{Loss Function for tattaka's Model}
When training this model, it is also necessary to address the imbalance issue in the heatmap, just as in the training of yu4u's model.
Here, we tackle this issue by using a slightly different loss function. Specifically, we compute the MSE loss separately for the positive and negative regions, then take the sum of their weighted means as the final loss function:
\begin{equation}
\begin{aligned}
    \mathcal{L}_\text{pos}(p, y) &= \frac{\sum (\text{MSE}(p, y) \cdot y)}{\sum y + \epsilon}, \\
    \mathcal{L}_\text{neg}(p, y) &= \frac{\sum (\text{MSE}(p, y) \cdot (1 - y))}{\sum (1 - y) + \epsilon}, \\
    \mathcal{L}_\text{tattaka}(p, y) &= \mathcal{L}_\text{pos}(p, y) + \mathcal{L}_\text{neg}(p, y),
\end{aligned}
\end{equation}
where $\epsilon$ is a small constant added to prevent division by zero.
This loss function not only resolves the imbalance issue but also accelerates convergence during training.
The hyperparameters used for training tattaka's model are shown in Table~\ref{tab:hyperparams2}.

\begin{table}[tb]
    \centering
    \footnotesize
    \begin{tabular}{ll}
        \toprule
        Training Epochs & 25 \\
        Learning Rate & $10^{-3}$ \\
        Optimizer & AdamW \\
        Weight Decay & $10^{-2}$ \\
        Warmup Epochs & 5 \\
        LR Scheduling Strategy & Cosine Decay \\
        Batch Size & 32 \\
        EMA Decay & 0.999 \\
        \bottomrule
    \end{tabular}
    \caption{Hyperparameters used for training tattaka's model.}
    \label{tab:hyperparams2}
\end{table}

\subsection{Inference Procedure}
Finally, we used four yu4u's models and three tattaka's models in the final submission.
To complete prediction within the time limit, we optimized our models by converting them to TensorRT format for faster inference. The conversion process was based on the notebook~\cite{Lion2025}.  
Additionally, we selected a Kaggle Notebook instance with dual T4 GPUs and leveraged multiprocessing to parallelize inference.

\begin{figure}[tb]
    \centering
    \includegraphics[width=0.25\textwidth]{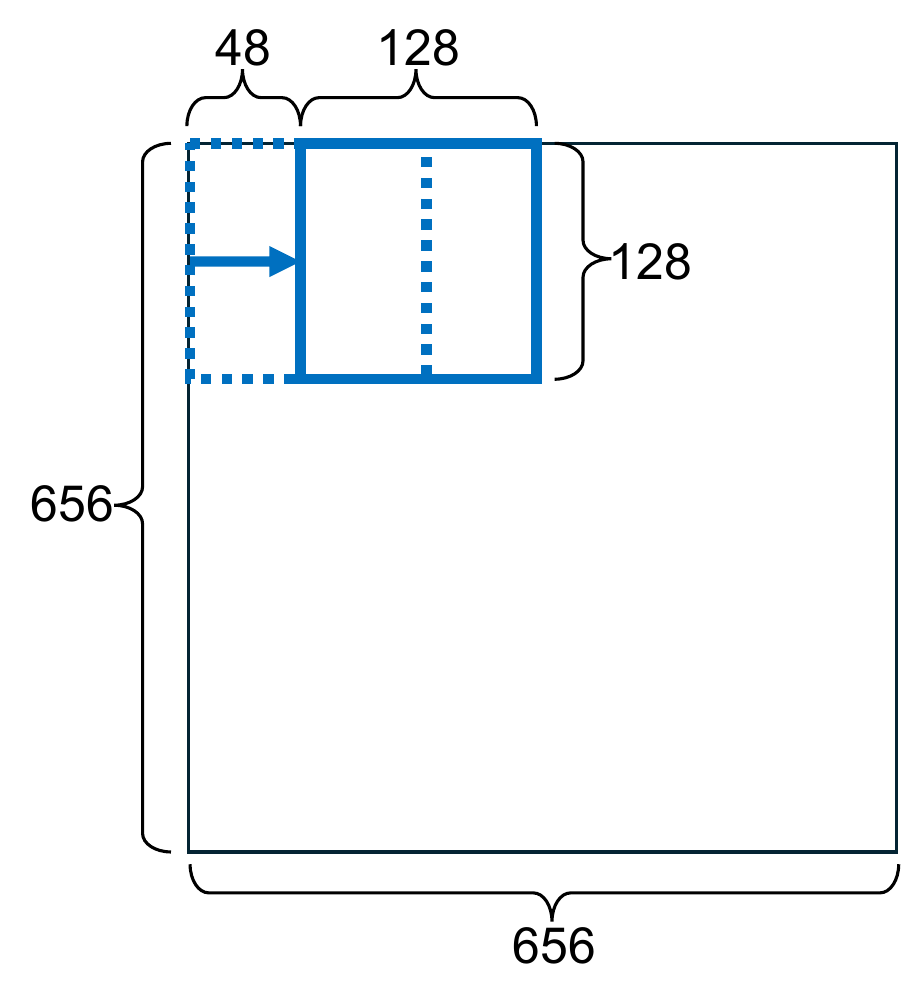}
    \caption{Sliding overlapping windows used in inference.}
    \label{fig:window}
\end{figure}

The input size in the XY dimensions for both models is 128×128, while the target inference size is 630×630. Therefore, inference is performed by sliding overlapping windows as shown in Fig.~\ref{fig:window}.
First, the input is padded to 656×656, and inference is conducted by moving a 128×128 window with a stride of 48. In this case, the number of inference windows becomes 12×12.
In the z-direction, yu4u's model, which has an input depth of 16, moves with a stride of 8, while tattaka's model, which has an input depth of 32, moves with a stride of 16 during inference.
The results of these models are all aggregated by taking the average.

Since the values near the edges of the window in the inference results have lower confidence, we apply weighting to reduce their impact on the final result.
This is achieved by creating a weight matrix of the same size as the inference result, where the center value is 1 and the weights decrease linearly to 0 at the edges of the window.

\subsection{Post Processing}
For the final heatmap, we first detect local maxima using Non-Maximum Suppression (NMS), which is efficiently implemented via max pooling with a kernel size of 7. The detected points are then filtered using different thresholds for each particle type.

Since the detected points are in the pixel coordinate system, we need to convert them to the particle coordinate system.
This is done as follows.
\begin{enumerate}[label=\arabic*.]
  \item \textbf{Centering}: Add 0.5 to the pixel coordinates to shift from the pixel’s top-left to its center.
  \item \textbf{Offset Correction}: Subtract the 1.0 offset that was added during heatmap generation.
  \item \textbf{Scaling}: Multiply by 10.012 to convert the adjusted pixel coordinates to the particle coordinate system.
\end{enumerate}

\section{Results}
Table~\ref{tab:results} shows the competition results of our approach, including scores on the public leaderboard and private leaderboard.
Our final submission is an ensemble of four yu4u's models with different folds and three tattaka's models.

Table~\ref{tab:results} also presents two versions of our submission: one where the number of windows in the XY direction was reduced from $12\times12$ to $8\times8$, and another where no weighting was applied based on the location within the window during inference.  
As the results indicate, using more windows --- i.e., increasing the overlap between windows --- and applying location-based weighting within windows during inference are both crucial for improving the score.

\begin{table}[tb]
    \centering
    \small
    \begin{tabular}{lcc}
        \toprule
        Method & Private LB & Public LB \\
        \midrule
        final submission & 0.783 & 0.788 \\
        $8 \times 8$ window & 0.779 & 0.784 \\
        w/o weight & 0.776 & 0.780 \\
        \bottomrule
    \end{tabular}
    \caption{Final submission performance with ensemble models on the competition leaderboards.}
    \label{tab:results}
\end{table}

\subsection{Things That Does Not Work}
Below, we present representative approaches that our team attempted but did not perform well.

\begin{itemize}
  \item \textbf{Two-stage model}: We built a model that refines the scores by cropping regions around the points detected using a heatmap approach and then applying a classification model to those cropped regions. Although it worked well in terms of CV scores, it did not improve LB performance. This may be due to the high difficulty of appropriately adjusting the threshold for the first stage and the second stage in the case of a two-stage model.
  \item \textbf{Detection model}: We built a CenterNet-like~\cite{Duan2019} object detection model (more precisely, a point detection model), but it achieved a significantly lower CV score compared to heatmap-based methods.
\end{itemize}

\section{Conclusion}
This paper introduced the detailed approach of the yu4u \& tattaka team in tackling the CZII - CryoET Object Identification competition.
Our solution adopted a heatmap-based keypoint detection approach, utilizing an ensemble of two different types of 2.5D U-Net models with depth reduction.
Despite its highly unified and simple architecture, our method achieved 4th place, demonstrating its effectiveness. We hope that our approach will contribute to the advancement of machine learning-based recognition of cryoET tomograms.

{
    %\small
    \footnotesize
    \bibliographystyle{ieee}
    \bibliography{main}
}


\end{document}